\documentclass[a4paper,conference]{IEEEtran}
\usepackage{amsmath}
\usepackage{graphicx}
\usepackage{subcaption}
\usepackage{diagbox}
\usepackage{latexsym}

\ifCLASSINFOpdf
\else
\fi
\begin{document}
%
\title{MSFD:Multi-Scale Receptive Field Face Detector}

\author{\IEEEauthorblockN{Qiushan Guo}
\IEEEauthorblockA{Beijing University of \\
Posts and Telecommunications\\
Beijing, China \\
Email: qsguo@bupt.edu.cn}\\

\and
\IEEEauthorblockN{Yuan Dong and Yu Guo}
\IEEEauthorblockA{Beijing University of \\
Posts and Telecommunications\\
Beijing, China \\
Email: \{yuandong, guoyu24k\}@bupt.edu.cn}\\

\and
\IEEEauthorblockN{Hongliang Bai}
\IEEEauthorblockA{Beijing Faceall Technology Co.,Ltd\\
Beijing, China \\
Email: hongliang.bai@faceall.cn}}


%


\maketitle

\begin{abstract}
We aim to study the multi-scale receptive fields of a single convolutional neural network to detect faces of varied scales. This paper presents our Multi-Scale Receptive Field Face Detector (MSFD), which has superior performance on detecting faces at different scales and enjoys real-time inference speed. MSFD agglomerates context and texture by hierarchical structure. More additional information and rich receptive field bring significant improvement but generate marginal time consumption. We simultaneously propose an anchor assignment strategy which can cover faces with a wide range of scales to improve the recall rate of small faces and rotated faces. To reduce the false positive rate, we train our detector with focal loss which keeps the easy samples from overwhelming. As a result, MSFD reaches superior results on the FDDB, Pascal-Faces and WIDER FACE datasets, and can run at 31 FPS on GPU for VGA-resolution images.
\end{abstract}


%
\IEEEpeerreviewmaketitle

\section{Introduction}
As one of the fundamental problems in computer vision and pattern recognition, face detection is the key step of various tasks like face alignment \cite{zhu2016face}, face recognition \cite{sun2014deep} and expression analysis. Face detectors of earlier stage are based on hand-crafted features like Viola-Jones \cite{viola2004robust}. Those hand-crafted detectors fail to handle complex problems in practical applications such as varied scale of faces, illumination conditions, various poses and facial expressions, etc.

In recent years, CNN-based methods have made great progress in classification and detection tasks. Considering the excellent representation power of CNNs, many elegant object detectors have appeared. In general, face detection can be viewed as a special case for generic object detection. Many of the face detectors with excellent performance trail the ideas of anchor-based detection methods like RCNN\cite{Girshick_2015_ICCV} and SSDs \cite{liu2016ssd}. These methods regress a series of anchors with pre-set shape towards objects and classify them. However, subtle changes are needed to handle face detection. We take three typical aspects to illustrate.

\subsection{Structure}
RCNN-based methods put anchor on the last layer of the CNN. But when the object becomes small, the recall rate will drop dramatically, so with the small faces. Higher-level features with larger receptive fields have an abstract semantic information, tending to ignore small ones. To get better result, low-level features containing rich feature together with suitable receptive field are needed. Therefore, we utilize feature pyramid for our face detection framework.

\subsection{Feature fusion strategy}
The Single Shot Detector (SSD) \cite{liu2016ssd} is one of the first attempts at using a ConvNet’s pyramidal feature hierarchy. Lately, FPN network \cite{lin2016feature} achieves a top-down information transmission making it possible for abstract semantic information to be transmitted to low-level features, thus providing context information for the detection of small objects. Our fusion of features exploits the context relationship between features on layers of different levels. It is also worth noticing that the feature fusion also brings changes in the receptive field, and we only fuse adjacent features to avoid too large receptive fields, which barely make contribution to the detection of small faces.


\subsection{Class imbalance}
In order to improve the recall rate of faces, anchors should be dense or matching threshold should be loose. However, densely assigning anchors like S$^{3}$FD \cite{zhang2017s} produces a pretty large number of negative examples, leading to extreme unbalance sample distribution. A common solution is to perform some form of hard negative mining by selecting hard negative ones to feed into training or carrying out more complex sampling/reweighting schemes. In this work, we refer to focal loss \cite{DBLP:conf/iccv/LinGGHD17} to handle the class imbalance and train efficiently on examples without easy negatives overwhelming the loss. 

The main contributions of this paper can be summarized as:
\begin{itemize}
\item We propose a novel feature agglomeration framework for face detection, which gets more context and texture from the adjacent feature maps so as to increase recall rate.
\item We introduce an anchor assignment strategy to improve the recall rate of rotated faces and outer faces.
\item We adopt focal loss to deal with the imbalance over face and background and reduce the high false positive rate of small faces. And we prove that focal loss can improve recall rate of face.
\item We achieve superior results on PASCAL-Faces, FDDB and WIDER FACE with less cost of computation compared with the state-of-the-art performances.
\end{itemize}

\begin{figure}[htbp]
\begin{center}
\includegraphics[width= \columnwidth]{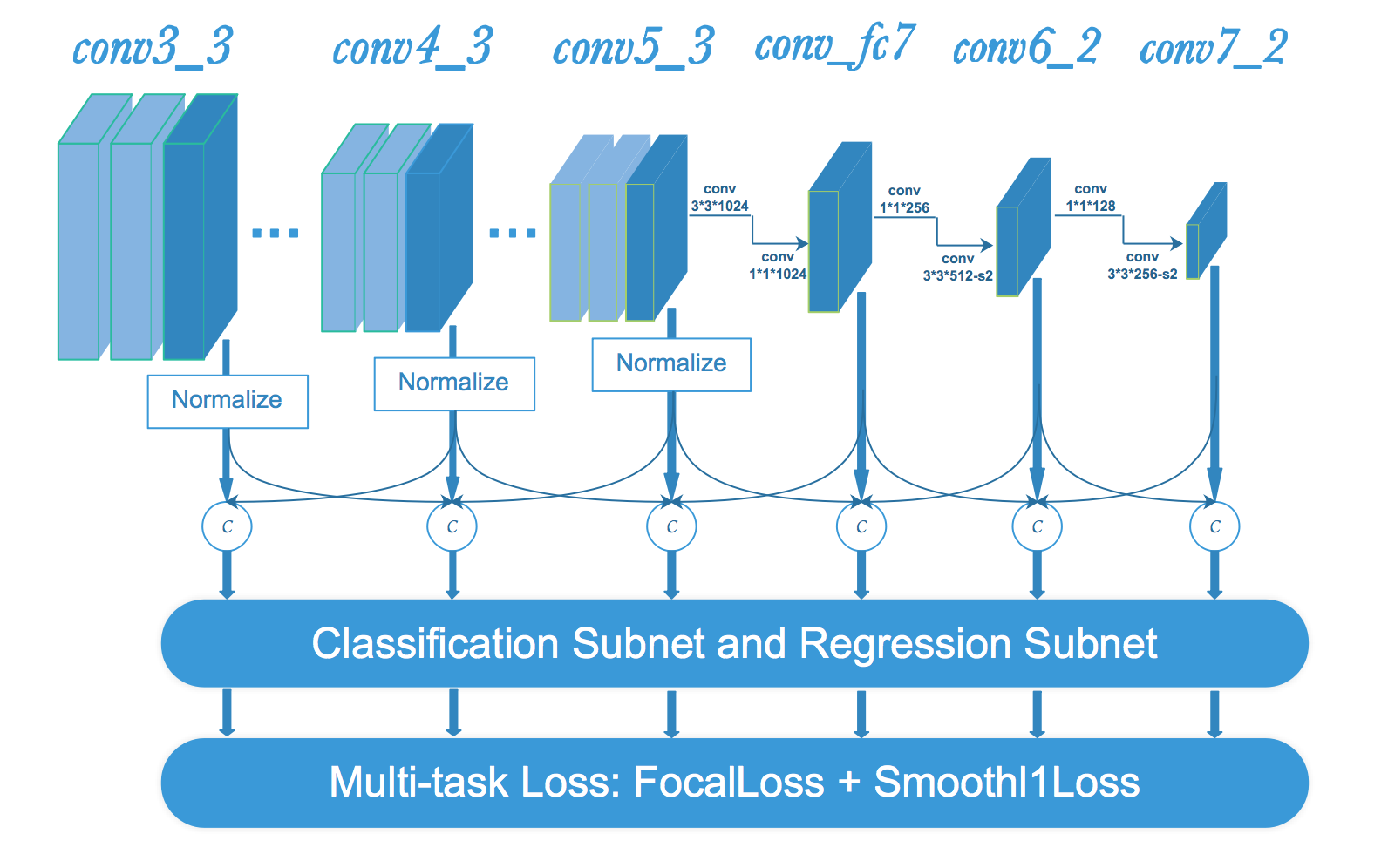}
\caption{The network architecture of Multi-Scale Receptive Field Face Detector. We apply L2 normalization to rescale conv3\_3, conv4\_3 and conv5\_3.}
\label{fig:architecture}
\end{center}
\end{figure}

\section{Related Work}

As a fundamental literature in the computer vision, face detection has been extensively studied in recent years. Viola-Jones detection framework \cite{viola2004robust} is a groundbreaking work using Harr feature and Adaboost to train a cascade classifier, achieving a fairly good result. Since then, researchers have focused on designing more powerful hand-crafted features \cite{li2014efficient,liao2016fast,yang2014aggregate,chen2014joint,zhu2012face,mathias2014face} to improve the performance. However, those traditional approaches rely heavily on the effectiveness of hand-crafted feature and optimize solely on each component, causing a sub-optimal problem to the whole pipeline.

In recent years, as the deep learning techniques, especially the convolutional neural networks(CNNs), gradually gain popularity and produce remarkable results on numerous computer vision tasks, CNN-based face detectors has become the mainstream. Among these, CascadeCNN \cite{li2015convolutional} and MTCNN \cite{zhang2016joint} both train a cascade structure for detection, while the latter uses multi-task CNNs to solve detection and alignment jointly. Yang \emph{et al}. \cite{yang2015facial} trains multiple CNNs for facial attributes to enhance the detection of occluded faces. 

Naturally, face detection can be regarded as a special case for generic object detection, the framework of which may also be transfered to fit the face detection task. Faster R-CNN \cite{ren2015faster} is one of the state-of-the-art detection pipelines composing of two stages. Based on that, Jiang \emph{et al}. \cite{jiang2017face} build a face detector and the performance is fairly good. Wan \emph{et al}. \cite{wan2016bootstrapping} and Sun \emph{et al}. \cite{sun2017face} both add some effective stategies including hard example mining and feature fusion on Faster R-CNN in order to achieve better results, while CMS-RCNN \cite{zhu2017cms} attaches body contextual information as well. What's more, Wang \emph{et al}. \cite{wang2017detecting} adopt another two-stage framework, R-FCN, to build their detector and  state-of-the-art on FDDB dataset \cite{jain2010fddb}. 

\begin{figure}[!htbp]
\begin{center}
\includegraphics[width= .5\textwidth]{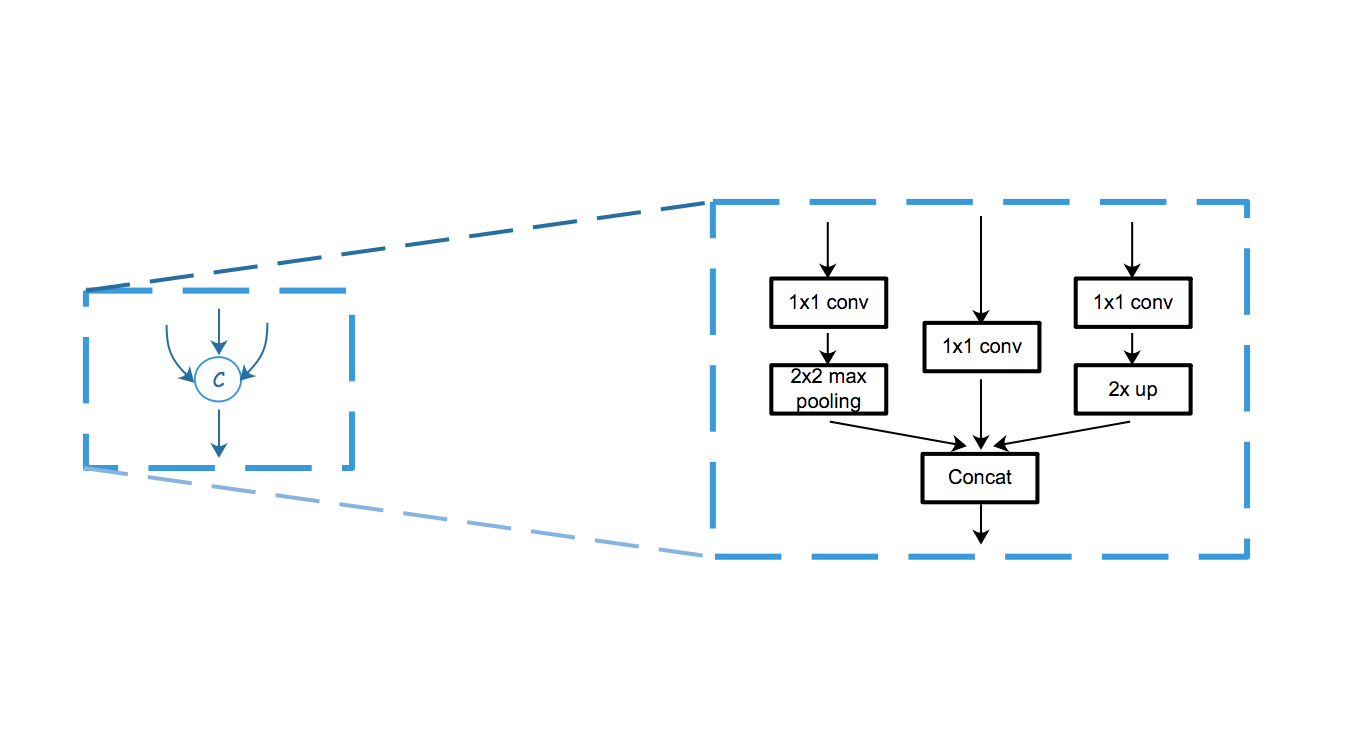}
\caption{The Context-Texture module. Bi-linear interpolation is employed to enlarge the size of deeper feature map by two times and $2\times2$ max-pooling to diminish shallower one.}
\label{fig:context}
\end{center}
\end{figure}

The single-stage detector including SSD \cite{liu2016ssd} and YOLO \cite{redmon2016you} is another popular form of detection pipeline which simultaneously performs classification and regression. SSH \cite{najibi2017ssh} is the typical single stage detector with context modules. Inspired by SSD and RPN \cite{ren2015faster}, Zhang \emph{et al}. propose S$^{3}$FD \cite{zhang2017s} with anchor matching strategy and max-out background label to ensure state-of-the-art performance on WIDER FACE \cite{yang2016wider} with real-time speed. In this work, we develop a superior face detector with real-time speed which adopts pyramidal feature hierarchy and aggregates multi-scale features with Context-Texture module.

\section{method}

\subsection{General Architecture}
Our goal is to leverage a ConvNet’s pyramidal feature hierarchy, which has more semantics at high levels and more texture at low levels. To this end, we propose a novel hierarchical feature agglomeration structure which aggregates adjacent features to increase recall rate.

\subsubsection{Constructing architecture}
The general architecture is shown in Fig.\ref{fig:architecture}. We inherit the backbone from \cite{zhang2017s} (based on the VGG16 network) and extract feature from conv3\_3, conv4\_3, conv5\_3, conv\_fc7, conv6\_2 and conv7\_2. They have the stride of \{4,8,16,32,64,128\} pixels with respect to the input image. Deeper layers have larger receptive field which can help to detect faces with different sizes. Most faces in the pictures from the Internet can be detected with the help of flexible receptive fields. As features from conv3\_3, conv4\_3 and conv5\_3 have different scales compared with the other layers', we apply L2 normalization \cite{DBLP:journals/corr/LiuRB15} to rescale them. And then we concatenate them with other features after transformation of the feature maps with different shapes.

\subsubsection{Context-Texture module}
To get more context and texture, we utilize Context-Texture block for adjacent selected feature maps with different shapes. The n-th selected layer feature is denoted as $f_{n}$ and the merged feature is denoted as $f^{'}_{n}$. The process of merging can be expressed as follows:
\begin{equation}
f^{'}_{n} = \mathcal{C}(f_{n-1},f_{n},f_{n+1}),
\end{equation}

\begin{figure}[!htbp]
\begin{center}
\includegraphics[width= .5\textwidth]{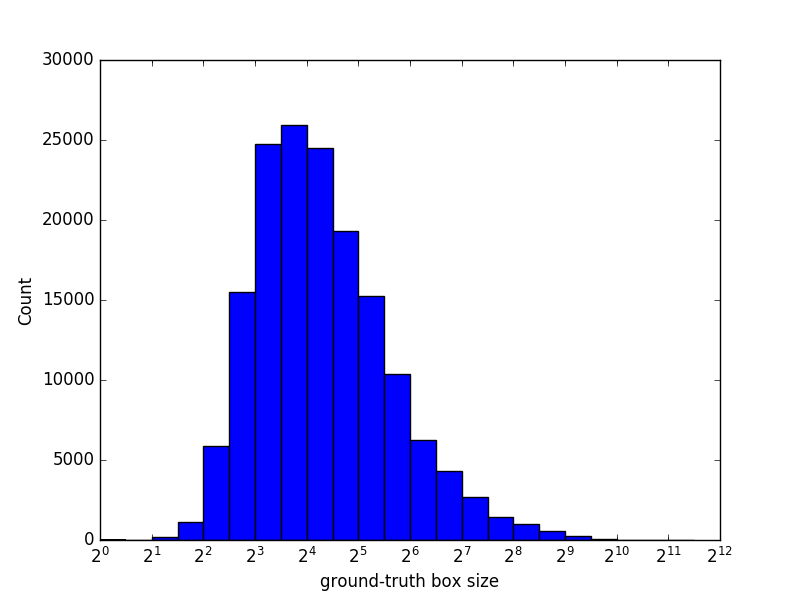}
\caption{The distribution of face scale in the WIDER FACE train set.}
\label{fig:distribution}
\end{center}
\end{figure}

where $\mathcal{C}$ is the Context-Texture module in Fig. \ref{fig:context}. The function's input doesn't contain $f_{n-1}$ when $n=1$. And the same goes for $f_{n+1}$ when $n+1$ is greater than the sum of the number of feature maps. Each Context-Texture block agglomerates the shallower feature $f_{n-1}$ and the deeper feature $f_{n+1}$ to enrich receptive field.

However, additional information is helpful but should not be overwhelming. Supposing that the channel of $f_{n}$ is reduced to $N$(e.g. 256) by using $1\times1$ convolutional filters, we reduce the channels of $f_{n-1}$ and $f_{n+1}$ to $\frac{N}{8}$ in the same way. After that, we reshape feature maps from different levels to the same 2-D shape by adding bi-linear interpolation to deeper ones and max-pooling with stride = 2 to shallower ones. 
The final agglomerative feature $f^{'}_{n}$ is obtained by concatenating these three features. Considering that too large or too small receptive fields can degrade the performance of detection \cite{hu2017finding} as well as time and memory consumption, we only take adjacent features for agglomeration.

\subsubsection{Classification Subnet and Box Regression Subnet}
To the agglomerated feature map attaches two subnetwork, one for classifying the anchor boxes and the other one for regressing from anchor boxes to ground-truth boxes. 
The classification subnet applies two $3\times3$ conv layers, each with 128 filters and followed by ReLU activations. The conv layers are then fed forward to a $3\times3$ conv layer with $K\times A$ filters where $K$ is the number of classes and $A$ is the number of anchors per location.
These two $3\times3$ conv layers can extract semantic information from the agglomerative feature map to classify accurately.
In parallel with the classification subnet, the box regression subnet is the same with the classification subnet except that the last conv layer outputs $4\times{A}$ relative offsets between the anchor and the ground-truth box.

\begin{figure}[htbp]
\begin{center}
\includegraphics[width= .5\textwidth]{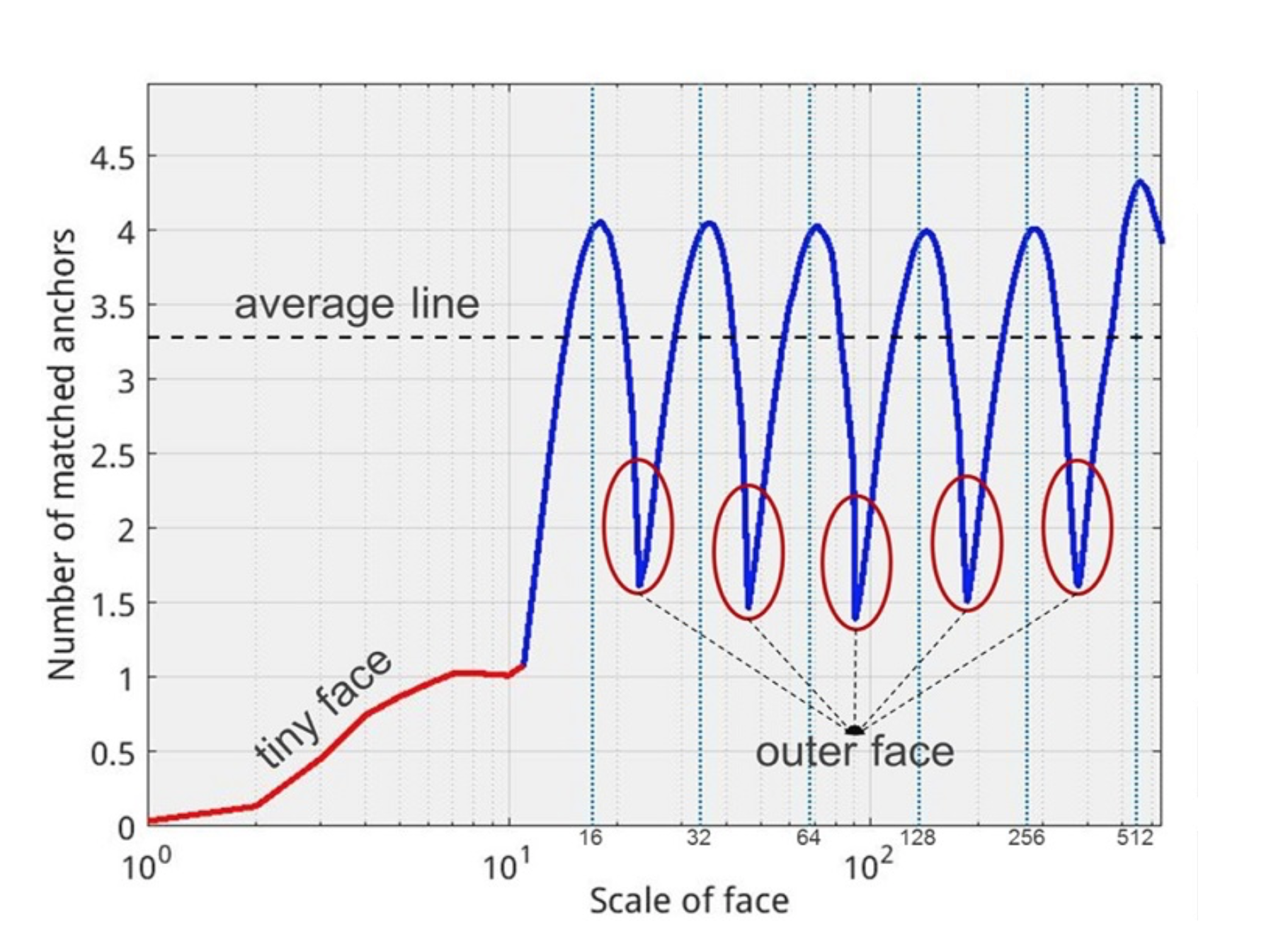}
\caption{This figure shows the number of anchors matched by faces of different sizes. The tiny and outer faces match less anchors. The figure is from \cite{zhang2017s}}
\label{fig:outer}
\end{center}
\end{figure}

\subsection{Anchor Assignment Strategy}
During training, we need to determine which anchor corresponds to a face bounding box. Anchor box can match faces whose size is similar. However, the size of the face ranges widely. As shown in Fig. \ref{fig:distribution}, most faces have an object size from 4 to 362 pixels. So, we define the anchors to have areas of \{$16^{2}$,$32^{2}$,$64^{2}$,$128^{2}$,$256^{2}$,$512^{2}$\} on these agglomerative feature maps. Each scale corresponds to a feature layer. Faces whose scale distribution lies away from anchors' scale can not match enough anchors, such as tiny and outer face in Fig. \ref{fig:outer}, leading to their low recall rate. Increasing the density of anchors and reducing the difficulty of matching can both contribute to a higher recall rate. To increase the density of anchors, we set the anchors' aspect ratios to 1 and 1.5 depending on the mutable aspect ratios of faces. Anchors are assigned to a ground-truth box with the highest IoU larger than 0.5, and to background if the highest IoU is less than 0.4. Unassigned anchors are ignored during training. 

\subsection{Training}
\subsubsection{Loss function}
Face detection methods encounter a salient class imbalance during training. We apply this model to a large number of random images and find that about 99.9\% of the anchors belong to negative samples and only a few of them are positive ones. In order to solve the extreme imbalance of positive and negative samples, most face detection frameworks adopt online hard negative mining strategy, which is helpful. However, the recently proposed focal loss \cite{DBLP:conf/iccv/LinGGHD17} is more powerful. We employ a multi-task loss function to jointly optimize model parameters:
\begin{equation}
\begin{split}
L(\{p_{i},t_{i}\})=&\frac{\lambda}{N_{cls}}\sum_{i}L_{cls}(p_{i},p^{*}_{i})+\\ &\frac{1}{N_{reg}}\sum_{i}I(p^{*}_{i}=1)L_{reg}(t_{i},t^{*}_{i}),
\end{split}
\end{equation}
where i is the index of an anchor and $p_{i}$ is the predicted probability of whether anchor $i$ is a face. The ground-truth label $p^{*}_{i}$ is 1 if the anchor is positive, 0 for negative. As defined in \cite{ren2015faster}, $t_{i}$ is a vector representing the 4 parameterized coordinates of the predicted boundingbox and $t^{*}_{i}$ is that of the ground-truth box associated with a positive anchor. The classification loss $L_{cls}(p_{i},p^{*}_{i})$ is focal loss over two classes (face and background) parameterized with $\alpha=0.25$ and $\gamma=2$. The regression loss $L_{reg}(t_{i},t^{*}_{i})$ is smooth L1 loss defined in \cite{Girshick_2015_ICCV}. $I(p^{*}_{i}=1)$ is the indicator function that limits the regression loss only focusing on the positively assigned anchors. The two losses are balanced by $\lambda$. $N_{cls}$ and $N_{reg}$ are the number of the positive anchors.

\subsubsection{Training dataset and data augmentation}
Our model is trained on $12,880$ images of the WIDER FACE training set. The distribution of face scale for this set is shown in Fig. \ref{fig:distribution}. The faces with size below 20 pixels affect the average precision of hard detection tasks in WIDER FACE. So, we randomly crop square patches with scale ranging from 0.3 to 1 of the shorter side from original image for training. In addition, the overlapped part of the face box is discarded if its center is out of the sampled patch. After randomly cropping, we employ color distortion strategy to preprocess training images, e.g. the adjustment of brightness, contrast, and saturation. Finally, the square patch is resized to $640\times640$ and horizontally flipped with probability of 0.5.

\subsubsection{Other implementation details}
We use pre-trained \cite{russakovsky2015imagenet} VGG16 as backbone. The parameters of conv\_fc6 and conv\_fc7 are initialized by subsampling parameters from fc6 and fc7 of VGG16 and the other additional layers (conv6\_1, conv6\_2, conv7\_1, conv7\_2) are randomly initialized with the “Xavier” method \cite{glorot2010understanding}. We initiate all convolutional layers except the final one for the classification subnet with bias $b=0$ and a Gaussian weight filler with $\sigma=0.01$. The final convolutional layer is initiated with bias $b=-log((1-\pi)/\pi)$ and $\pi=0.01$ here. And the $\lambda$ in Eq.(2) is set to 3 to balance the loss of classification and regression. We use SGD optimizer with momentum of 0.9, weight decay of 0.0005, and a total batch size of 16 on 4 GPUs. The maximum number of iterations is $120k$ and the learning rate starts at $10^{-3}$ and becomes 10 times smaller at $80k$ and $100k$ iterations. Our implementation is based on Caffe \cite{jia2014caffe}.

\section{Experiments}

\subsection{Model analysis}
We analyze our model on the WIDER FACE validation set by extensive experiments and ablation studies. According to the difficulty of detection tasks, the validation set is set split to easy, medium and hard subsets. The evaluation metric is mean average precision (mAP) with Interception-of-Union (IoU) threshold of 0.5. 

\subsubsection{Baseline}
To evaluate our contributions, we adopt the closely related detector S$^{3}$FD as the baseline. This is because both our method and S$^{3}$FD base on the same backbone. We compare our method with the performance of the S$^{3}$FD under two different settings: (i) $S^{3}FD(F)$: it only uses the scale-equitable framework. (ii) $S^{3}FD$($F$+$S$+$M$): it is the complete model. Here, $S$ is the scale compensation anchor matching strategy which makes faces match enough anchors and $M$ is the the max-out background label to address the unbalanced binary classification problem. For optimization and training, S$^{3}$FD adopts online hard negative mining with softmax loss.

\begin{table*}[!htbp]
\centering
\begin{tabular}{p{4cm}<{\centering} || p{2cm}<{\centering} | p{2cm}<{\centering} | p{2cm}<{\centering} | p{2cm}<{\centering} | p{2cm}<{\centering}}
\hline
Method&S$^{3}$FD(F)&MSFD(C)&MSFD(C+F)&S$^{3}$FD(F+S+M)&  MSFD  \\ 
\hline
Scale compensation + Max-out& & & &$\surd$& \\
\hline
Context-Texture Module& &$\surd$&$\surd$& &$\surd$\\
\hline
Focal Loss& & &$\surd$& &$\surd$\\
\hline
Anchor Assignment& & & & &$\surd$\\
\hline
Easy (mAP \%)&92.6&93.8&94.6&93.7&94.7\\
\hline
Medium (mAP \%)&91.6&92.7&93.5&92.5&93.5\\
\hline
Hard (mAP \%)&82.3&83.8&85.0&85.2&86.5\\
\hline
\end{tabular}
\caption{Ablation studies of MSFD. All settings are trained on the training set of WIDER FACE and then tested on the validation set. MSFD(C) adopts the same implementation details with S$^{3}$FD(F) except Context-Texture Module. MSFD(C+F) is trained with focal loss. MSFD is the complete model with anchor assignment}
\label{tab:compare}
\end{table*}

\begin{figure*}[h!]
  \centering
  \begin{subfigure}[b]{.3\textwidth}
    \includegraphics[width=\textwidth]{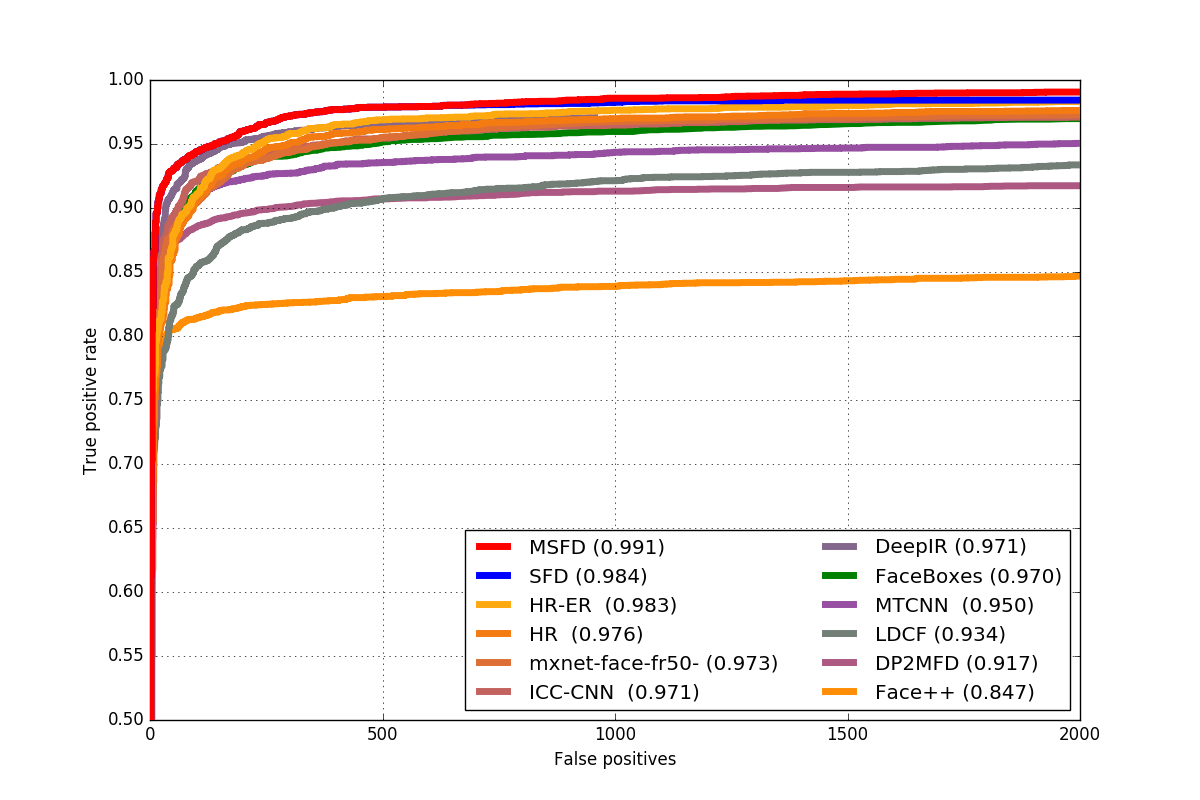}
    \caption{FDDB discontinuous ROC curves.}
    \label{fig:fddbdROC}
  \end{subfigure}
  \begin{subfigure}[b]{.3\textwidth}
    \includegraphics[width=\textwidth]{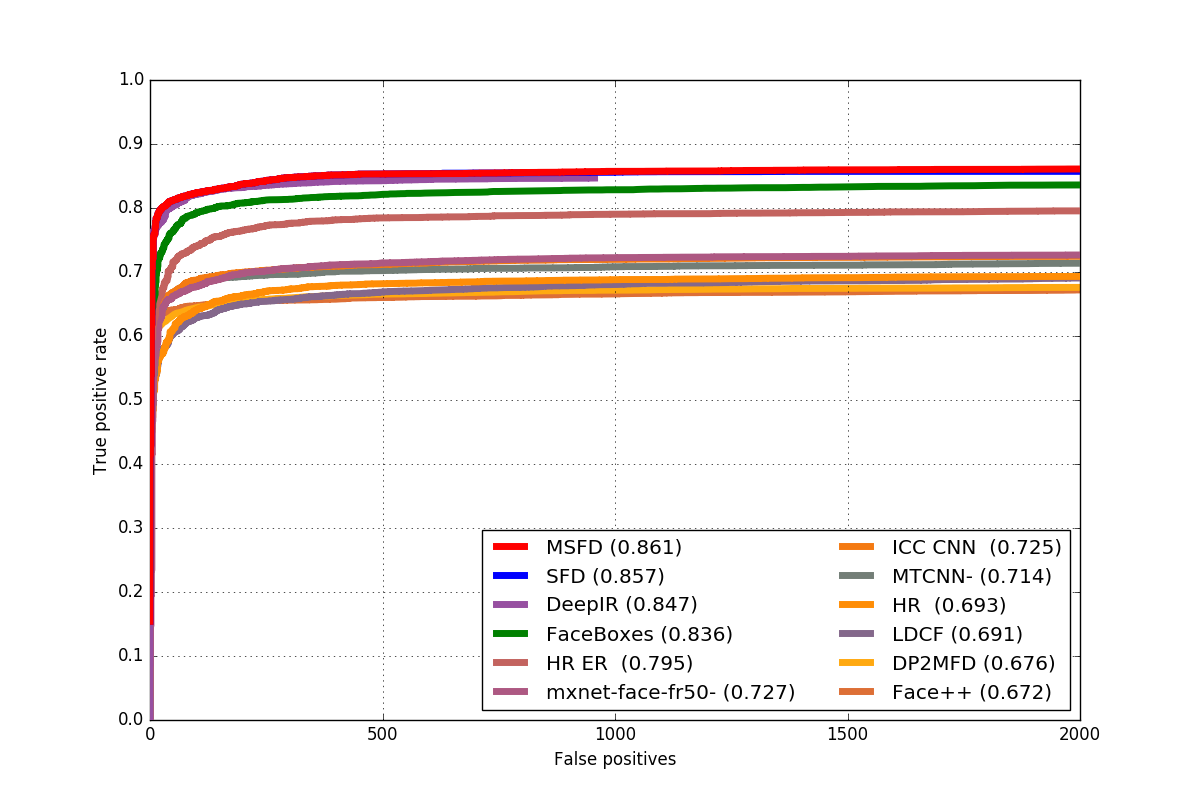}
    \caption{FDDB continuous ROC curves.}
    \label{fig:fddbcROC}
  \end{subfigure}
  \begin{subfigure}[b]{.3\textwidth}
    \includegraphics[width=\textwidth]{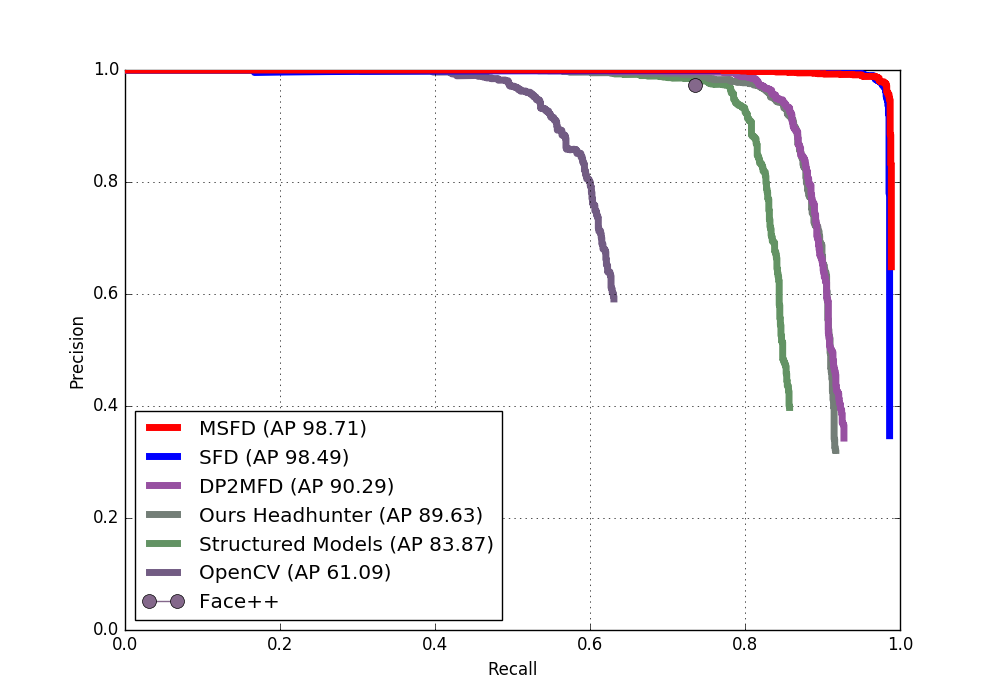}
    \caption{Pascal Faces ROC curves.}
    \label{fig:pascalROC}
  \end{subfigure}
  \caption{Evaluation on the FDDB and PASCAL Face dataset}
  \label{fig:ROC}
\end{figure*}

\begin{figure*}[h!]
  \centering
  \begin{subfigure}[b]{.3\textwidth}
    \includegraphics[width=\textwidth]{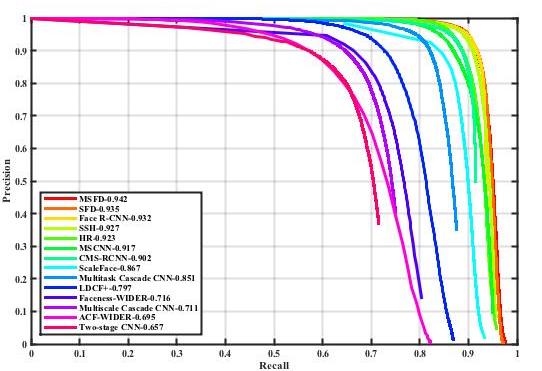}
    \caption{Easy.}
    \label{fig:val_easy}
  \end{subfigure}
  \begin{subfigure}[b]{.3\textwidth}
    \includegraphics[width=\textwidth]{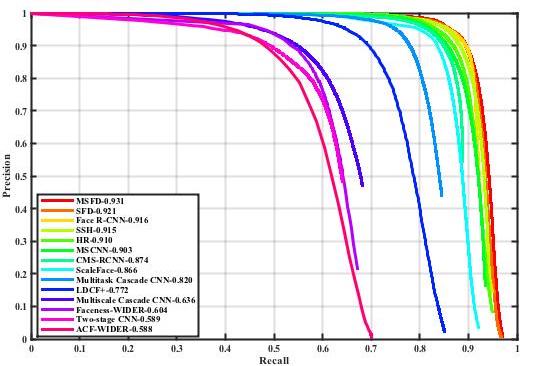}
    \caption{Medium.}
    \label{fig:val_medium}
  \end{subfigure}
  \begin{subfigure}[b]{.3\textwidth}
    \includegraphics[width=\textwidth]{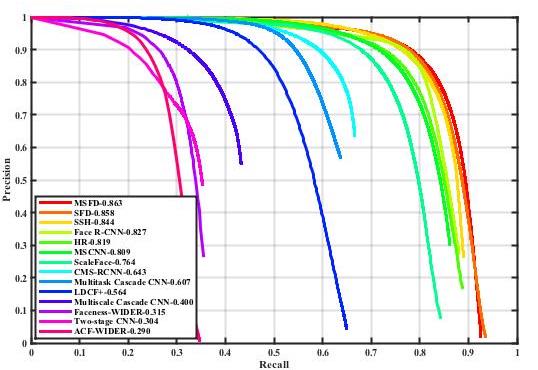}
    \caption{Hard.}
    \label{fig:val_hard}
  \end{subfigure}
  \caption{Precision-recall curves on WIDER FACE test set. S$^{3}$FD updated the results, whose mAP of hard subset is increased by 1.8\% compared with the paper \cite{zhang2017s}, on the WIDER FACE test set.}
  \label{fig:val_roc}
\end{figure*}

From the results listed in TABLE \ref{tab:compare}, some conclusions can be summed up in the next subsections.
\subsubsection{Context-Texture module can help us judge more accurately}
Context-Texture module agglomerates adjacent features in order to enrich the receptive fields. To better understand the impact of feature agglomerating, we adopt the same anchor assignment strategy, training parameters and implementation details with $S^{3}FD(F)$. TABLE \ref{tab:compare} indicates that the performance gains significant improvement.

\subsubsection{Focal loss is elegant for handling the class imbalance}
The second and third column in TABLE \ref{tab:compare} show that focal loss can effectively solve the problem of sample imbalance over face and background. The mAP of easy, medium and hard subset is increased by 0.8\%, 0.8\%, 1.2\%. The increases mainly come from preventing the vast number of easy negatives from overwhelming the detector during training.

\subsubsection{Anchor assignment strategy can match more}
The result shows that our anchor assignment strategy can make faces in hard subset match enough anchors so that the mAP of hard subset improve 1.5\%. This strategy makes the rotated and tiny faces, which is the main component of hard subset, match more anchors. But increasing the density of large anchors seems less helpful to the results of easy and medium subset.

\subsection{Evaluation on benchmark}
We evaluate our MSFD method on the common face detection benchmarks, including PASCAL-Face \cite{yan2014face}, FDDB \cite{jain2010fddb} and WIDER FACE \cite{yang2016wider}.

\subsubsection{FDDB dataset}
The dataset is a well-known benchmark and it contains 5,171 faces in 2,845 images. We transform the predicted bounding boxes to ellipses to get a precise result. We adopt annotations released from \cite{zhang2017s} to avoid false positive faces with high scores caused by unlabelled faces. Fig. \ref{fig:fddbdROC} and Fig. \ref{fig:fddbcROC} show the evalution results, our method achieves promising results on both discontinuous and continuous ROC curves compared with the previous state-of-the-art methods \cite{zhang2017s,zhang2016joint,hu2017finding,Zhang_2017_ICCV,sun2017face,DBLP:journals/corr/abs-1708-05234,DBLP:journals/corr/Ohn-BarT17,ranjan2015deep}. This indicates that our MSFD can detect unconstrained faces robustly.

\subsubsection{PASCAL-Face dataset}
This dataset was collected from PASCAL person layout test subset. It has 1,335 labeled faces in 851 images. Fig. \ref{fig:pascalROC} shows the precision-recall curves. Our method gets 98.71\% mAP which outperforms the previous state-of-the-art detector S$^{3}$FD (98.49\%) and SSH (98.27\%)\cite{najibi2017ssh}, and beats the other methods \cite{ranjan2015deep,mathias2014face,yan2014face,viola2004robust}.

\subsubsection{WIDER FACE dataset}
It contains 32,203 images with 393,703 annotated faces with a high degree of variability in scale, pose and occlusion. The training set has 158,989 faces and these are 40\% of the total set. The validation set accounts for 10\% and the test set accounts for 50\%. According to the difficulty of detection tasks, the validation and test set are is devided into ``easy", ``medium" and ``hard" subsets. The difficulty is determined by the size of the face in the picture. Most faces in ``hard'' subset have a small shape. We train our detector on the training set and test on both validation and test set. 
The precision-recall curves and mAP are shown in Fig. \ref{fig:val_roc}. It achieves 94.2\%(Easy), 93.1\%(Medium) and 86.3\%(Hard) for test set. And 94.7\%(Easy), 93.5\%(Medium) and 86.5\%(Hard) for validation set. 
This outperforms overwhelming majority of the submitted results.


\subsection{Inference time}
Our method outputs lots of boxes, so we should better filter out the boxes with low confidence by a threshold of 0.05 and keep the top 300 boxes before NMS. Then we apply NMS with jaccard overlap of 0.3 and keep the top 200 boxes. We test our computation cost on NIVIDIA 1080Ti with Intel Xeon E5-2670v2@2.50GHz. For the VGA-resolution image, our detector can run at 31 FPS and achieve the real-time speed. The majority of the time consumption is spent on the VGG16 backbone network, so a lightweight network could be more efficient.

\section{Conclusion}
This paper proposes a novel face detector by enriching the receptive field with Context-Texture module. The proposed method has superior performance on various common face detection benchmarks and enjoys real-time inference speed on GPU. We analyze the relationship between receptive field and the accuracy of face detection module, then propose a method to agglomerate contextual information and textural information by hierarchical structure. Moreover, we propose the dense anchor assignment strategy to improve the recall rate of small faces and outer faces. And we train the model robustly in an end-to-end manner with focal loss to deal with the large class imbalance over small faces and background. The experiments demonstrate that our method results in the superior performance. In future work, we intend to further improve the anchor assignment strategy. It's crucial to generate more accurate anchors to reduce the cost of computation and false positives.

\section{Acknowledgment}
This work is supported by Chinese National Natural Science Foundation under Grants 61532018.






%

{\small
\bibliographystyle{plain}
\bibliography{gqs_ref}
}

\end{document}